\documentclass{article}
\usepackage{spconf,amsmath,graphicx}
\usepackage[colorlinks]{hyperref}
\usepackage{url}

\title{Image-Based Virtual Try-on System With Clothing-Size Adjustment}
\name{Minoru Kuribayashi, Koki Nakai, Nobuo Funabiki
\thanks{This study was supported by the JSPS KAKENHI (22K19777), JST SICORP (JPMJSC20C3), JST CREST (JPMJCR20D3), and ROIS NII Open Collaborative Research (2022-22S1402), Japan.}
}
\address{
Graduate School of Natural Science and Technology\\
Okayama University, Okayama, Japan.
}
\begin{document}
%
\maketitle
\begin{abstract}
The conventional image-based virtual try-on method cannot generate fitting images that correspond to the clothing size because the system cannot accurately reflect the body information of a person. In this study, an image-based virtual try-on system that could adjust the clothing size was proposed. The size information of the person and clothing were used as the input for the proposed method to visualize the fitting of various clothing sizes in a virtual space. First, the distance between the shoulder width and height of the clothing in the person image is calculated based on the coordinate information of the key points detected by OpenPose. Then, the system changes the size of only the clothing area of the segmentation map, whose layout is estimated using the size of the person measured in the person image based on the ratio of the person and clothing sizes. If the size of the clothing area increases during the drawing, the details in the collar and overlapping areas are corrected to improve visual appearance.
\end{abstract}

\section{Introduction}
With the increase in online shopping, the online apparel shopping market is growing rapidly. Online shopping is a convenient method for purchasing items over the Internet without physically visiting a store. However, when purchasing clothes online, the customer cannot try out clothes to check if they fit. 

Han et al.~\cite{viton} first proposed the virtual try-on (VITON) method, which can generate visualization images for users trying out clothes online. Allowing users to virtually try on clothes to check for fitting not only improves their shopping experience and changes the manner of shopping for clothes but also reduces costs for retailers. Variants~\cite{Wang_ECCV18,FW-GAN,VTNFP,2d-tryon,Yang_CVPR20} of VITON have been developed for improving virtual visual appearance. To improve the resolution of the synthetic image on the virtual try-on, Han et al.~\cite{Clothflow} predicted the optical flow maps of clothes and desired clothing regions. 

Choi et al. proposed a novel high-resolution virtual try-on method called VITON-HD~\cite{viton-hd}, in which a novel clothing-agnostic person representation leveraging the pose information and the segmentation map is plotted. Furthermore, in the alignment-aware segment normalization introduced in VITON-HD, information irrelevant to the clothing texture in the misaligned regions is removed and semantic information is propagated throughout the network. Although VITON-HD can generate photorealistic fitting images, the clothing size is not considered. Even if clothes of various sizes are used, the same output is always generated because the system deforms clothes to fit the shape and design of a person's body, and the shape of the clothes to be changed is based on that of the clothes worn in the person's image. 

In this study, an image-based virtual fitting system that can adjust the clothing size was proposed. Because VITON-HD generates a segmentation map based on the clothes a person is wearing, a fitting image that differs from the clothing size is generated. In the proposed method, the cloth area of the segmentation map is changed according to the ratio of the actual person and cloth sizes by using the size information of the person and clothes as the input. The clothing size is changed according to the size of the person by calculating the distance between the shoulder width and height of the clothing in the person image. If only the cloth region is enlarged, then the boundaries of the cloth region, such as the collar and the overlap with an arm, may become inconsistent. Therefore, we changed the coordinates of the segmentation map and corrected the boundary of the cloth region to maintain quality.

\section{Virtual Try-On System}
Given an image of a desired cloth and that of a person, an image-based virtual fitting system generates an image of the person wearing this cloth. In this section, we briefly review image-based virtual fitting methods and explain drawbacks.

\subsection{VITON-HD}
VITON-HD~\cite{viton,viton-hd} consists of four independent modules, namely pre-processing, segmentation generation, clothes deformation, and try-on synthesis.

In this model, pose estimation and parts estimation of a person wearing the target clothes are used for virtual try-on image synthesis. First, the segmentation map and pose map are used to pre-process an input image as a clothing-agnostic person image and its corresponding segmentation.
Furthermore, semantic segmentation is used in VITON-HD to separate appearance and shape generation, which generates spatial and natural results. Thus, a clothing-independent representation of the person can be achieved, information about the cloth worn before fitting can be removed, and body information for reconstruction can be obtained. In skeletal estimation, parts of the body that are difficult to reconstruct (e.g., arms and hands) are removed.

The semantic segmentation generation module (SGM) generates masks for exposed body parts (i.e., composite body part masks) and warped clothing regions using semantic segmentation of body parts and clothing. The SGM has a UNet~\cite{U-net, spectral} structure consisting of a convolutional layer, a downsampling layer, and an upsampling layer. Furthermore, two multiscale discriminators~\cite{semantic-gans} are used for conditional adversarial loss. The SGM generates semantic masks in a two-step process---the body part is first generated and then the clothing mask is synthesized, which renders the original clothing shape in the person image completely network agnostic.
    
In clothes deformation, input clothes are deformed by a geometric transformation to fit a given person. A thin plate spline (TPS) transformation~\cite{tps} is then used to deform the clothing image, which is an extension of the spline curve to a two-dimensional plane. These parameters were used to warp the clothing image to fit the estimated layout.
    
The UNet-based network is used to generate virtual-fitting images using intermediate products (target layout, area of compositing process, and warped clothing images) obtained up to this key point as the input. The try-on synthesis module explicitly determines areas that cannot be complemented by the cloth warping and inputs them to the generation module to reduce image quality degradation resulting from misalignment.

\subsection{Problem}
Conventional image-based virtual fitting systems depend on the clothing size a person is originally wearing. In VITON-HD, a segmentation map is generated from an image of a person by using face, arms, legs, and clothes worn by the person as feature points. The segmentation map is then used as an input for geometric transformation processing, which generates a virtual try-on image according to the size of the clothes worn by the person before try-on.

If the clothing region indicated by a segmentation map is extended, then nonclothing areas are affected and the clothing area is enlarged. However, in this module, appearance cannot be changed to a specified size.

\section{Proposed Method}
In the proposed method, the clothing size is changed to match the actual size by adding real-space size information and using a correction process to VITON-HD segmentation.

\subsection{Overview}
\begin{figure}
    \centering
    \begin{minipage}{3cm}
        \centering
        \includegraphics[scale=0.15]{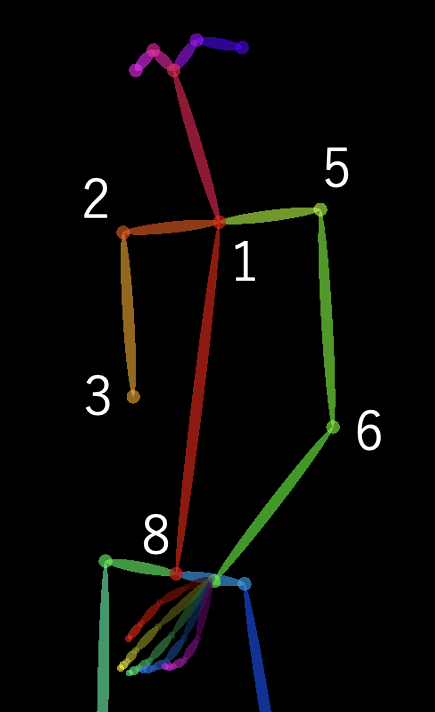}
        \caption{Some key points detected by OpenPose.}
        \label{fig:keypoint}
    \end{minipage}
    \hspace*{5mm}
    \begin{minipage}{4cm}
        \centering
        \includegraphics[scale=0.25]{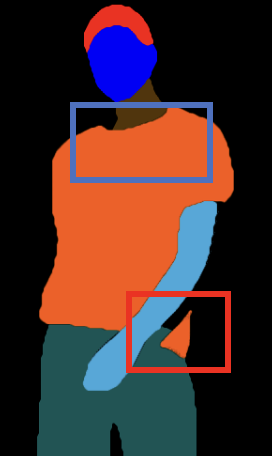}
        \caption{Irregularities at the collar region and overlap with the clothing area.}
        \label{fig:irregular}
    \end{minipage}
\end{figure}

To change the clothes size according to the person, the actual sizes of the person and clothes were added as inputs to the correction process. Size information is defined as the shoulder width in the horizontal direction and the height of clothes in the vertical direction.
Virtual space information is calculated based on the coordinates of the key points detected using OpenPose~\cite{openpose,openpose2} and the distance between the shoulder width and height of the person image. Figure~\ref{fig:keypoint} displays these details. 

Only the clothing region of the segmentation map is resized, the layout of which is estimated using the size of the person measured in the person image based on the ratio of the actual person size to the clothing size. By changing the clothing size ratio to match the person in the person image based on the size ratio between the actual person and the clothing, the clothing size in the person image is changed to the clothing size corresponding to the person.

Changing the clothing size results in the coordinates of the segmentation map to differ from those of VITON-HD, which causes inconsistencies at the boundaries. Therefore, a correction technique was applied to the boundary of the clothing region to maintain the same quality as that of VITON-HD. 

\subsection{Ratio to clothes Size}
To reflect the clothing size in the person image, the actual clothing and person size information were first added as inputs. Size information is obtained from the shoulder width and height of the clothes in the horizontal and vertical directions, respectively. In OpenPose, the clothing size can be applied to the person image according to the input size information.

\subsubsection{Adjustment of the clothing size}
To reflect the clothing size in the person image, the clothing and person size information are first added as inputs. Size information is calculated from the shoulder width and height of the clothes in the horizontal and vertical directions, respectively.

The vertical size of the clothing in a person image is obtained by detecting 25 key points by using OpenPose and calculating the ratio of the coordinate distance between key points 1 and 8, the size of the actual clothes, and the height of the person. Suppose that the height of a person's clothing region is $H_p$, the height of the clothing is $H_c$, and the coordinate of key point $t$ is $(x_t, y_t)$. The clothing’s height $\tilde{H}$ in a virtual space can be obtained using the following equation:
\begin{equation}
\tilde{H} = \frac{H_c}{H_p} \times \delta_{1,8},
\end{equation}
where $\delta_{i,j}$ is the Euclidean distance between key points $i$ and $j$ is defined as follows:
\begin{equation}
\delta_{i,j} = \sqrt{(x_i-x_j)^{2} + (y_i-y_j)^{2}}.
\end{equation}

Similarly, the horizontal size $\tilde{W}$ of the clothing in a virtual space can be calculated using the following equation:
\begin{equation}
\tilde{W} = \frac{W_c}{W_p} \times \delta_{2,5},
\end{equation}
where $W_c$ and $W_p$ are the shoulder widths of the clothing and person, respectively.

The lateral size of the clothing was calculated from the distance between key points 2 and 5 in the figure, the distance between key points 3 and 2, and the distance between key points 5 and 6. In this case, (sleeve length + shoulder width) $\alpha$ in the person’s image can be calculated using the following equation:
\begin{equation}
    \alpha = \delta_{2,5} + \delta_{2,3} + \delta_{5,6}
\end{equation}

From the segmentation map obtained by layout estimation, only the clothes regions were obtained by color extraction. Using $\tilde{H}$, $\tilde{W}$, and $\alpha$, the clothing area was enlarged according to the vertical and horizontal distances to reflect the size of the clothes to be changed.

\subsection{Adjustment of Details}
As displayed in Fig.~\ref{fig:irregular}, when clothes are enlarged, some irregularities may occur. To suppress such side effects, the following two post-operations were performed using the proposed method.

\subsubsection{Collar Region}
When the segmentation map was merged by enlarging the clothes region, a gap appeared around the collar because the size of the segmentation map created by VITON-HD differed from that of the original shoulder coordinates when the clothes region was enlarged, and the other segmentation maps were combined.

In the proposed method, an erosion operator~\cite{book} was applied in morphological image processing to the collar area of the clothes region to generate a natural fitting image. Because clothes shrink if the contraction is applied to the entire clothes region, limiting the contraction to the collar region is necessary. Therefore, we consider a rectangle whose center is key point 1 in Fig.~\ref{fig:collar}, whose horizontal direction $s_x$ is proportional to the distance between key points 2 and 5, and whose vertical direction $s_y$ is proportional to the cloth’s size. In this study, we use $s_x=3\tilde{W}/4$ and $s_y=3\tilde{H}/4$.

\subsubsection{Overlapping Arms in Clothes Area}
If an arm overlaps the clothing region, as displayed in the red frame in Fig.~\ref{fig:irregular}, the positions of the clothes and arm regions become misaligned because the clothes region is divided and the coordinates of only the clothes region change when enlarged, whereas the coordinates of the arm region do not. Therefore, the coordinates of the shortest distance between the clothes and arm regions were obtained before and after enlargement, and the coordinates of the clothes region were corrected while maintaining the shortest distance between the two regions based on the coordinates of the clothes region.

\begin{figure}
    \centering
    \begin{minipage}{3cm}
        \centering
        \includegraphics[scale=0.45]{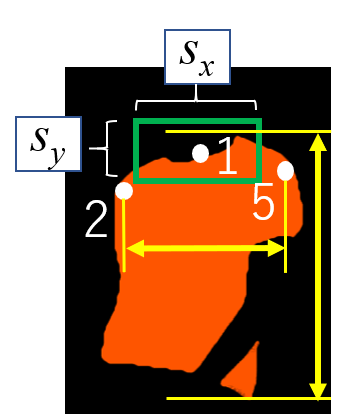}
        \caption{Correction of the collar region.}
        \label{fig:collar}
    \end{minipage}
    \hspace*{5mm}
    \begin{minipage}{4.5cm}
        \centering
        \includegraphics[scale=0.24]{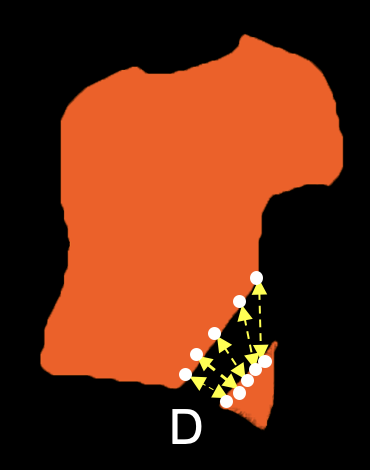}
        \caption{Shortest distance between two separated regions.}
        \label{fig:contour}
    \end{minipage}
\end{figure}

Before enlargement, the distance between the two clothes areas is calculated for some combinations of the path in the two separated regions, and the minimum distance $D$ is determined, as displayed in Fig.~\ref{fig:contour}. When the two clothes regions are enlarged from the center coordinates, the coordinates with the shortest distance are calculated. The coordinates with the shortest distance are then obtained by finding all the distances between the two regions by the combination of the path and obtaining the coordinate with the minimum value.

The difference between the coordinates of the shortest distance before enlargement and the coordinates of the shortest distance after enlargement is determined, and the center coordinates of each object is moved by the difference in the shortest distance. Let $(x_C, y_C)$ denote the center coordinates of regions, and let $(x, y)$ and $(x^\prime, y^\prime)$ denote the coordinates before and after enlargement, respectively. By using the differences in the horizontal $\Delta_x=x^\prime_C-x_C$ and vertical $\Delta_y=y^\prime_C-y_C$ directions before and after enlargement, the smaller region is shifted to a larger region. By moving the center coordinates of the two clothing regions, the clothes region can be expanded while maintaining the distance between the arm regions.

\section{Evaluation}
To evaluate whether the proposed method can perform virtual fitting corresponding to the sizes of actual clothes, the person’s images, person, and clothes sizes were measured.

\begin{figure}
    \centering
    \includegraphics[scale=0.6]{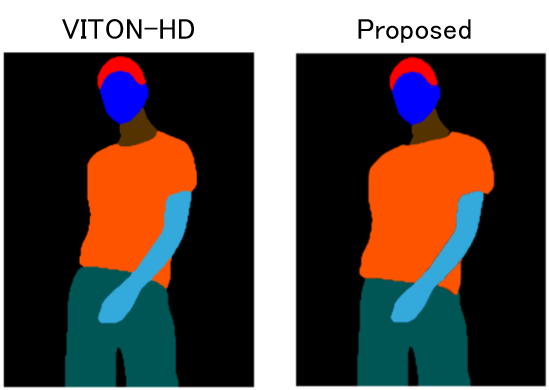}
    \caption{Comparison of the generated segmentation map.}
    \label{fig:segmentation}
\end{figure}

The segmentation map generated in VITON-HD~\cite{viton-hd} and the proposed method are displayed in Fig.~\ref{fig:segmentation}. The segmentation map of the proposed method is enlarged based on information of the shoulder width and height of the person and clothing. In the collar area, the clothes area can be enlarged while maintaining the same height as the person’s shoulder. Even when the clothes area overlaps with the arm area, the distance between the arm area and the clothes area can be enlarged without shifting the position of the arm area and the clothes area.

\begin{figure}
    \centering
    \includegraphics[scale=0.61]{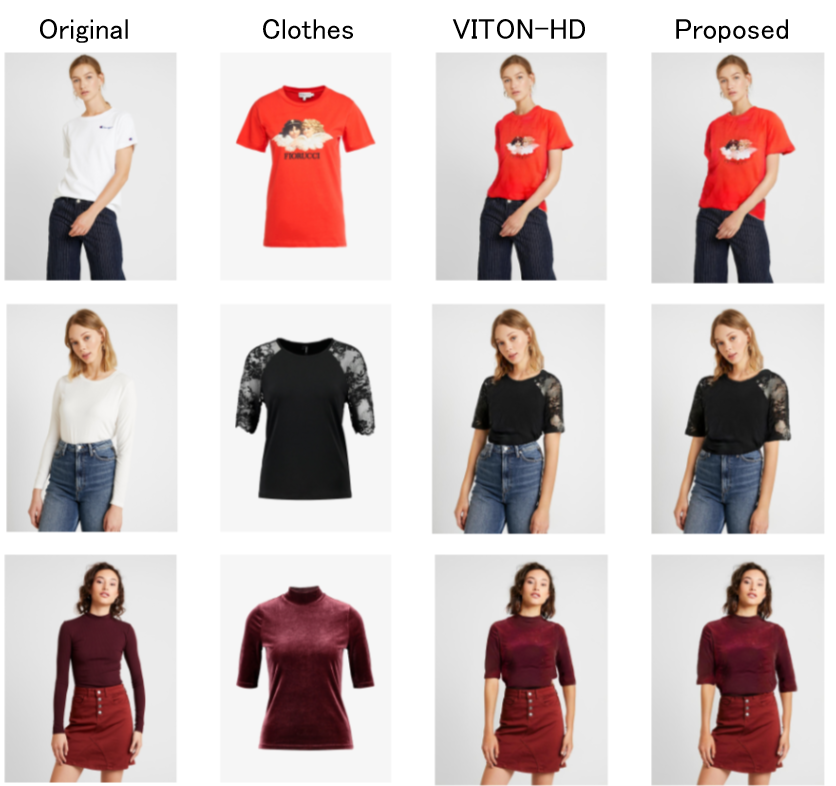}
    \caption{Comparison of generated images.}
    \label{fig:example}
\end{figure}

The generated fitted images with VITON-HD using the segmentation map created by the proposed method are presented in Fig.~\ref{fig:example}. The body and clothes length information was set to 66 and 73 cm, respectively, and the clothes size varied considerably. In VITON-HD, the fitting image is generated according to the size of the clothing before changing, and the size of the clothing to be changed is not changed, whereas the proposed method changes the clothing area in the fitting image according to the size of the clothes. Even when an arm overlapped the clothing area, the proposed method generated a fitting image of the appropriate size without separating it from the arm. 

To confirm the effectiveness of the proposed method, a subjective evaluation experiment was conducted to evaluate whether the proposed method exhibits superior performance. Four testers (A, B, C, and D) tried on clothes and the shoulder width and height of the clothes, and ten students in our faculty responded to the questionnaire. The number of students participating in the proposed method are summarized in Table~\ref{tab:result}.
To maintain fair evaluation, the order of VITON-HD and the proposed methods was randomly changed, and the proposed method was not revealed. The results of the questionnaire revealed that the proposed method is more natural and appropriate for resizing than VITION-HD is.

\begin{table}[t]
\centering
  \caption{Results of the questionnaire from ten users, where the number of users who chose the proposed method is displayed.}
  \label{tab:result}
  \begin{tabular}{|c|c||c|c|c|c|c|} \hline
   \multicolumn{2}{|c||}{} & \multicolumn{4}{|c|}{tester} & \\\cline{3-6}
   \multicolumn{2}{|c||}{} & ~ A ~ & ~ B ~ & ~ C ~ & ~ D ~ & clothes \\ \hline \hline
  size & height & 52 & 45 & 46 & 38 & 62 \\ \cline{2-7}
  (cm) & width & 47 & 41 & 43 & 32 & 51 \\ \hline\hline
  \multicolumn{2}{|c||}{Results} & 10 & 8 & 7 & 10 & --- \\\hline
  \end{tabular}
\end{table}

\section{Conclusions}
In this study, an image-based virtual fitting system that corresponds with the clothing size was proposed, which applies OpenPose in the image segmentation region. The segmentation map generated by VITON-HD can be resized accurately by estimating the skeletal structure using OpenPose and adding clothing size, body length, and shoulder width as inputs.
In the future, appropriate size changes not only for T-shirts but also for various other types of clothing and the generation of natural-looking virtual fitting images according to the shooting environment will be studied.

\bibliographystyle{IEEEbib}
\bibliography{refs}

\end{document}